\pdfoutput=1

\documentclass[11pt]{article}

\usepackage[final]{acl}

\usepackage{times}
\usepackage{latexsym}

\usepackage{multirow}
\usepackage{booktabs}
\usepackage{amssymb}
\usepackage{amsmath}
\usepackage{subcaption}
\usepackage{graphicx}

\usepackage[T1]{fontenc}

\usepackage[utf8]{inputenc}

\usepackage{microtype}

\usepackage{inconsolata}

\title{SememeLM: A Sememe Knowledge Enhanced Method for Long-tail Relation Representation}

\author{Shuyi Li \\
 Tianjin University\\
  \texttt{Suyielee@tju.edu.cn} \\
   \And
  Shaojuan Wu \\
  Tianjin University \\
  \texttt{shaojuanwu@tju.edu.cn} \\
  \AND
  Xiaowang Zhang \\
  Tianjin University \\
  \texttt{xiaowangzhang@tju.edu.cn}
  \And 
  Zhiyong Feng \\
  Tianjin University \\
  \texttt{zyfeng@tju.edu.cn} 
}

\begin{document}
\maketitle

\begin{abstract}
Recognizing relations between two words is a fundamental task with the broad applications. 
Different from extracting relations from text, it is difficult to identify relations among words without their contexts. Especially for long-tail relations, it becomes more difficult due to inadequate semantic features. 
Existing approaches based on language models (LMs) utilize rich knowledge of LMs to enhance the semantic features of relations. 
However, they capture uncommon relations while overlooking less frequent but meaningful ones since knowledge of LMs seriously relies on trained data where often represents common relations. On the other hand, long-tail relations are often uncommon in training data.
It is interesting but not trivial to use external knowledge to enrich LMs due to collecting corpus containing long-tail relationships is hardly feasible.
In this paper, we propose a sememe knowledge enhanced method (SememeLM) to enhance the representation of long-tail relations, in which sememes can break the contextual constraints between wors. 
Firstly, we present a sememe relation graph and propose a graph encoding method. 
Moreover, since external knowledge base possibly consisting of massive irrelevant knowledge, the noise is introduced. We propose a consistency alignment module, which aligns the introduced knowledge with LMs, reduces the noise and integrates the knowledge into the language model. 
Finally, we conducted experiments on word analogy datasets, which evaluates the ability to distinguish relation representations subtle differences, including long-tail relations. 
Extensive experiments show that our approach outperforms some state-of-the-art methods.
\end{abstract}

\section{Introduction}

\begin{figure}[t]
 \centering
 \includegraphics[width=0.45\textwidth]{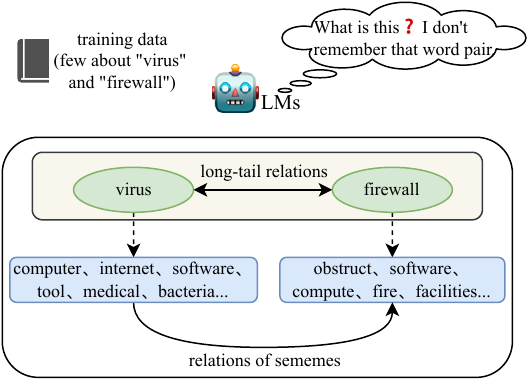}
 \caption{The long-tail relation problem and our solution}
 \label{intro}
\end{figure}

\begin{table*}[t]
    
    \caption{Two examples in E-KAR dataset, where the candidate in bold characters is the answer. The relations of word pairs are extracted from explanations provided by the dataset but are not provided when training and testing.} 
    \label{Tab}
    \begin{subtable}{0.35\linewidth}
      \centering
            \begin{tabular}{ll}
            \toprule
            Query: & ceramics: teacups \\             
            \midrule
            Cans: & \textbf{A) silver: necklace}  \\
            &B) clay: potted plants \\
            &C) goose feather: coat \\
            &D) steel: automotive \\ 
            \bottomrule
            \end{tabular}
    \end{subtable}%
    \begin{subtable}{0.65\linewidth}
      \centering
        \begin{tabular}{lll}
            \toprule
            Query: &   bee: honey & \textit{ProducedBy}, \textit{UseUp}\\
            \midrule
             Cans: & A) butteryfly: cocoon pupa & \textit{BecomeFrom}\\
             & B) hen: egg & \textit{ProducedBy}\\
             & C) father: child & \textit{FatherOf}\\
             & \textbf{D) farmer: food}  &  \textit{ProducedBy}, \textit{UseUp} \\
            \bottomrule
        \end{tabular}
    \end{subtable} 
\label{example}
\end{table*}

Relations hold a central position in a wide range of applications such as common sense question-answering, reading comprehension, knowledge graph completion, and other applications that depend on knowledge graphs \cite{DBLP:journals/isci/TangXZLCL23, DBLP:journals/pvldb/000123, DBLP:journals/jamia/PengYYBHW23, DBLP:conf/iclr/VashishthSNT20}. 
As LMs have advanced, extensive research has focused on extracting relations from context, leading to significant achievements. On the contrary, there remains a research gap in the representation of word pair relations in context-free scenarios. Especially for long-tail relations, the lack of semantic features makes relation recognizing more difficult.

A common strategy to model relations between word pairs is to take the vector difference between the representations of each word. But it has been shown that the relation vectors obtained through this method contain noise and its latent space only apply to some simple relations, such as \textit{capital-of} or \textit{singular-plural} \cite{DBLP:conf/ijcai/Murena22}. 
Another strategy is to utilize KGs to obtain relation representations, which is highly beneficial for acquiring simple lexical relations. However, they have certain limitations in terms of scalability because the relations provided by KGs are too coarse-grained. For example, the relation schema in KGs often fail to solve analogical reasoning problems or other relation similarity task. Table \ref{example} shows a instance of E-KAR analogy task. We may have access to triples in KGs such as (\textit{ceramics, MaterialOf, teacups}), but such knowledge is not sufficient to solve the given question, e.g. since all word pairs have relation \textit{MaterialOf}.   

Previous studies have shown that LMs store a large amount of factual knowledge \cite{bouraoui2020inducing, relbert, MultilingualKnowledgeTriples2022emnlp}, but they capture uncommon relations while overlooking less frequent but meaningful ones. The knowledge of language models heavily relies on trained data, which often represents common relations, while long-tail relations are typically underrepresented in the training data. Therefore, it is critical for using external knowledge to enrich LMs due to collecting corpus containing long-tail relationships is hardly feasible.

To this end, we propose a sememe knowledge enhanced method (SememeLM) to enhance the representation of long-tail relations, in which sememes can break the contextual constraints between wors. As shown in Figure \ref{intro}, words are composed of sememes, so the relations between words can be constructed from the relations between sememes. 
The most challenging aspect is the diversity of sememe combinations, which makes it difficult to obtain effective relations.
In response to this challenge, we obtain all sememes and their relations from OpenHowNet \cite{2003HowNet} to construct a sememe relation graph and propose a graph encoding method.
Since external knowledge base possibly consisting of massive irrelevant knowledge, the noise is introduced. To reduce the noise and integrate the sememe knowledge with LMs, we propose a consistency alignment module. 
To promote the model's focus on long-tail relations between word pairs, we leverage relation similarity data and incorporate supervised contrastive learning into our model training.
We conducted experiments on seven word analogy datasets. Extensive experiments show that our approach outperforms some state-of-the-art methods. 

The contribution of this paper can be summarized as follows: 
\begin{itemize}
\item We propose a sememe knowledge enhanced method (SememeLM) to improve long-tail relation representations and constructed a sememe relation graph. 
\item To integrate the graph and language models, we propose a consistency alignment module. This module aligns information at both the word representation and relation representation levels. 
\item We are the first to enhance relation representation using sememe knowledge and extensive experiments show that SememeLM outperforms some state-of-the-art methods. 
\end{itemize}


\section{Our approach}

\begin{figure*}[ht]
 \centering
 \includegraphics[width=\textwidth]{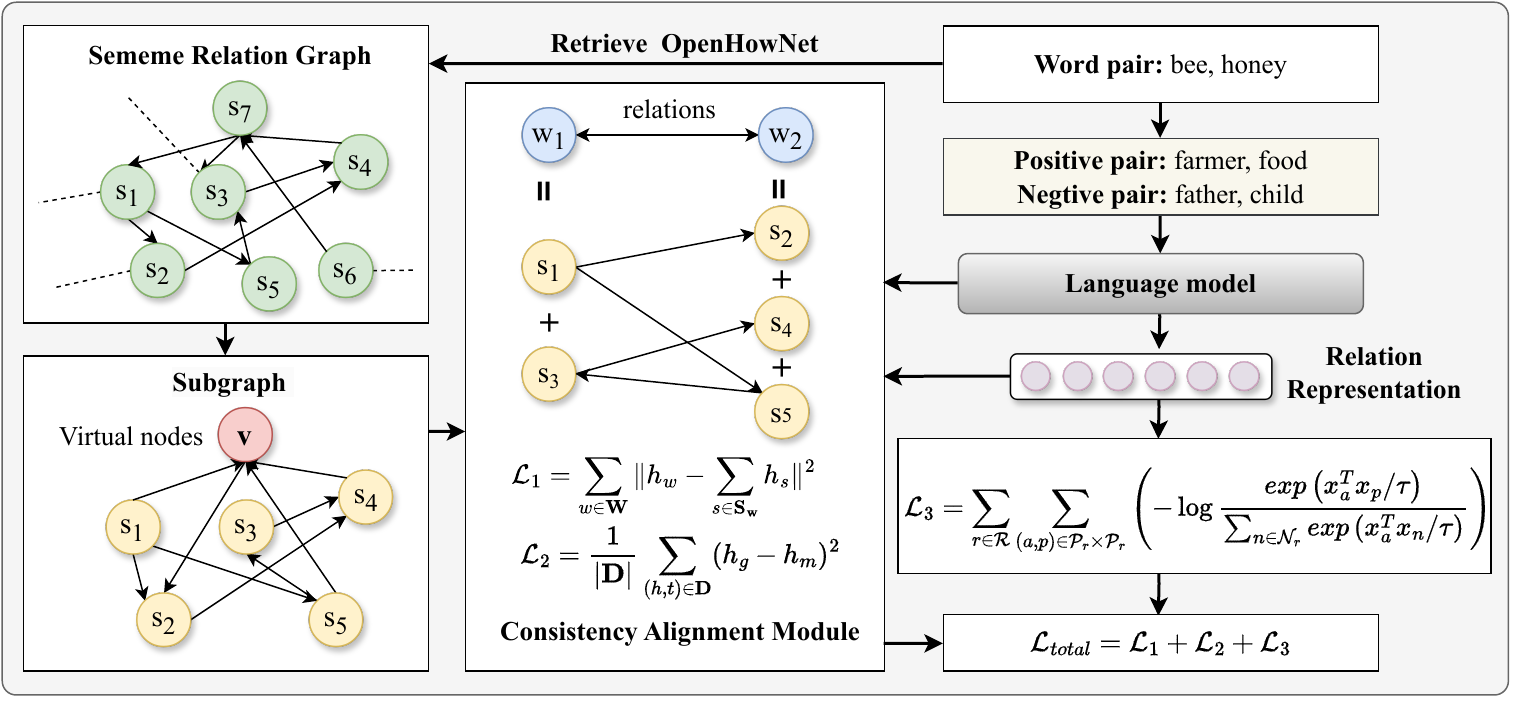}
 \caption{The overall flow of training our model when given word pairs.}
 \label{model}
\end{figure*}

In this section, we will provide a detailed explanation of our approach (SememeLM), whose framework is shown on Figure \ref{model}. In general, SememeLM is a relation representation model enhanced with sememe knowledge. The input of the model is word pairs, and the output is a representation of the relation between word pairs. First, we retrieved sememes and relations from HowNet and BabelNet to construct a sememe relation graph. 
Then we use a graph attention mechanism to learn representations on the graph. 
To integrate the KG and LM, we propose a consistency alignment module. 
Finally, we introduce supervised contrastive learning to train our model on relation similarity data.

\begin{figure}[ht]
 \centering
 \includegraphics[width=0.45\textwidth]{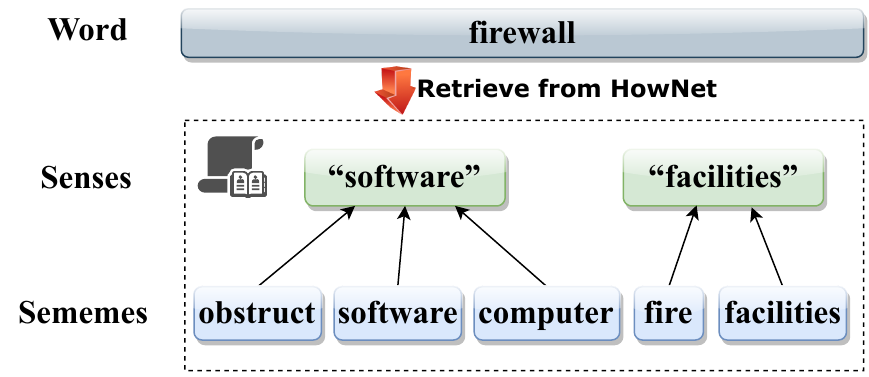}
 \caption{An example of the HowNet structure.}
 \label{hownet}
\end{figure}

\subsection{HowNet} \label{HowNetI}

HowNet \cite{2003HowNet} is an external knowledge base, the principle of which regards sememe as an atomic semantic unit. Each word in HowNet is assigned one or more senses, and each sense is associated with one or more sememes. 
Different from WordNet \cite{1994WordNet}, it offers minimal semantic of words which can facilitate the understanding and expression of semantic relations between words. 
It has been proven to improve word representation learning \cite{DBLP:conf/acl/NiuXLS17} and has been widely utilized in many NLP tasks such as text matching \cite{DBLP:conf/aaai/LyuCZ021}.

An example is illustrated in Figure \ref{hownet}. The meaning of the word "firewall" in the sense of "software" can be expressed by the combination of the three meanings of "software", "obstruct" and "computer", and in the sense of "facilities" it can be expressed by the combination of "facilities" and "fire". 

\subsection{Sememe relation Graph} 

We retrieve OpenHowNet \cite{2003HowNet} to obtain different sememes of each word and their relations, including 2,196 sememes and 403 different relations. 
The relations between sememes in OpenHowNet are derived from BabelNet\cite{babelnet}. 

Since sememes are typically part of common vocabulary in the corpus, the relations between sememes tend to be intuitive and common. We assume that relation representations constructed from combinations of sememes are the most basic units of relations. This helps us uncover meaningful but uncommon relations between word pairs. 

We traverse all the sememes and relations, ultimately obtaining a graph $\mathbf{G}=(\mathbf{V}, \mathbf{E})$. $\mathbf{V}$ is the set of nodes (sememes) and $\mathbf{E}$ is the set of edges (relations). For all pairs of sememes, if a relation exists between them, there is an edge representing their relation. It is worth noting that relations are directed, thus the graph is a directed graph.
   
For each word $w$, we denote the set of senses as $\mathbf{O}=\left\{o_1, o_2, \cdots, o_k\right\}$. 
$o_k$ is the k-th sense of $w$ and we denote its corresponding sememes as $\mathbf{S}_{i}=\left\{s_{1,i}, s_{2, i}, \cdots, s_{m, i}\right\}$. $s_{m, k}$ is the m-th sememe of the i-th sense $o_k$. 
As we concentrate on the sememes comprising words, we do not consider the word's sense. It's important to note that various senses may involve distinct sememes, and we consider their intersection:
\begin{equation}
     \mathbf{S}_w = \cap_{i=1}^{k} \mathbf{S}_{i}
\end{equation}
It's worth noting that if the intersection is empty, we do not encode the graph, and the model degenerates into a general pre-trained language model.
Since direct reasoning on the complete graph is intractable, we retrieve the corresponding sememes of the input word pair. And then we extract a subgraph $G=(\mathbf{V}, \mathbf{E})$ from the sememe relation graph. 

\subsection{Graph Encoding}
To encode the graph, we utilize a Graph Attention Network (GAT) \cite{veličković2018graph}. To acquire relation representations, we introduce a virtual node, with the initial representations of virtual nodes being randomly initialized. At the same time, some edges will be added to ensure that all sememes nodes point to the virtual node. The initial representations of sememe nodes are took from GloVe. 
 
At the l-th step, each node $v_i \in \mathbf{V}$ aggregates information by attending to its neighbors and itself. The weighted average of the connected nodes calculates the updated representation $h^l_i$ of $x_i$. The update of the node $v_i$ in GAT is as follows:
\begin{equation}
 \mathbf{h}_{i}^{l} =\sigma\left(\sum_{x_{j} \in \mathbf{N}^{+}\left(v_{i}\right)} \alpha_{i j}^{l} \cdot\left(\mathbf{W}^{l} \mathbf{h}_{j}^{l-1}\right)\right)
\end{equation}
where $\mathbf{W}^{l}$ is a learnable parameter, $N^+(v_i)$ is the set of neighboring nodes of $v_i$. and $ \sigma (\cdot) $ is a nonlinear activation function, e.g. LeakyReLU. 
The attention coefficient $\alpha_{i j}^{l}$ is the normalized correlation of the representation between the two nodes $v_i$ and $v_j$ in a unified space, i.e.

\begin{align}
\alpha_{i j}^{l} &= \operatorname{softmax} f^{l}\left(\mathbf{h}_{i}^{l-1}, \mathbf{h}_{j}^{l-1}\right) \\
 &=\operatorname{softmax}\left(\mathbf{W}_{q}^{l} \mathbf{h}_{i}^{l-1}\right)^{T}\left(\mathbf{W}_{k}^{l} \mathbf{h}_{j}^{l-1}\right)
\end{align}
where $\mathbf{W}_{q}^{l}$ and $\mathbf{W}_{k}^{l}$ are learnable parameters. The word representations of sememes $\textbf{h}_s$ will be obtained from the corresponding nodes. The relation representations of word pairs $\textbf{h}_g$ will be obtained from the virtual node. Since the representation on the graph is 300-dimensional, a matrix transformation is required to bring it into line with the dimensions of the pre-trained language model:
\begin{equation}
    \textbf{h}_g' = \textbf{W} \textbf{h}_g
\end{equation}
where $\textbf{W}$ is transformation matrix. 

\subsection{Consistency Alignment Module}
In order to obtain fine-grained relation representations from LMs, we utilize a template function to convert a given word pair into a sentence. Given a word pair $x_i = (h, t)$, it can an be expressed as $\mathcal{T}(h, t)$. 
We define $\mathcal{M}$ as a LM; therefore, encoding word pairs using a LM can be represented as follows:
\begin{equation}
    \mathbf{h}_m = \mathcal{M}(\mathcal{T}(x_i))
\end{equation}
Based on previous experience \cite{relbert}, we use the following template as "I finally discovered the relation between [h] and [t] : <mask>". We take the representation at masked positions as relation representations and use the representations at the respective word positions as word representations.

To integrate the KG and LM, we propose a consistency alignment module. The design of the consistency alignment module is based on the following intuitions: 
\begin{itemize}
    \item The words are composed of sememes, so the representation of a word is equivalent to the sum of its sememe representations.
    \item The relation representation between corresponding sememes of word pairs is equivalent to the relation between the word pairs.
\end{itemize}
Then we align word representations by minimizing:
\begin{equation}
    \mathcal{L}_1 = \sum_{w \in \mathbf{W}} \left \| \mathbf{h}_w  - \frac{1}{| \mathbf{S_w} |} \sum_{s \in \mathbf{S_w}} \mathbf{h}_s \right \| 
\end{equation}
where $\mathbf{W}$ represents the set of all words in the training dataset. 
In addition, we also align relation representations with MSE loss:
\begin{equation}
    \mathcal{L}_2 = \frac{1}{|\mathcal{D}|}  \sum_{(h, t) \in \mathcal{D}}  ( \mathbf{h}_g' - \mathbf{h}_m )^2
\end{equation}
where $\mathbf{D}$ represent the training dataset. 

\subsection{Training} \label{training}
We fine-tune the model using a supervised contrastive loss to distinguish fine-grained differences between words 
and we access to some positive training examples and a corresponding set of negative examples for a number of relations $\mathcal{N}_r$. 
$x_a$, $x_p$, $x_n$ are the corresponding relation representations. The model is trained to make the distance between $x_a$ and $x_p$ smaller than the distance between $x_a$ and $x_n$. The elements a, p and n correspond to word pairs, where $a, p, n \in \mathcal{P}$. Formally, this is accomplished using the following loss function:
\begin{equation}
    \mathcal{L}_3= \sum_{r \in \mathcal{R} } \sum_{(a,p) \in \mathcal{P}_r \times \mathcal{P}_r}  \left ( -\log{\frac{e
^{ {x_a^Tx_p}/{\tau} }}{ \sum_{n \in \mathcal{N}_r }  e^{ {x_a^Tx_n}/{\tau} }} }   \right ) 
\end{equation}
where $\tau$ is a temperature parameter to control the scale of the exponential. 

The total loss is the sum of word alingment loss $\mathcal{L}_1$, relation alignment loss $\mathcal{L}_2$ and supervised contrastive loss $\mathcal{L}_{3}$ calculated as follows:
\begin{equation}
 \mathcal{L}_{total} =  \mathcal{L}_{1} + \mathcal{L}_{2} + \mathcal{L}_{3}
\end{equation}


\section{Experiments and Results}
This section presents how models perform on relation representation. In Section \ref{4.1}, we first introduce the settings of experiments. Then, we approximately demonstrate the comparison results of SememeLM in Section \ref{4.2}. Section \ref{4.3} shows an ablation study of our proposed model. Finally, we provide some examples to analyze how our method is effective in Section \ref{4.4}.

\begin{table*}[ht]
    \centering
    \caption{We compare our method with small-sized models and human performance on different benchmarks.}
        \begin{tabular}{ccccccccc}
        \toprule[1.5pt]
        \textbf{Model}  & \textbf{BATS} & \textbf{UNIT 2} & \textbf{UNIT 4} & \textbf{Google} &\textbf{SAT} &  \textbf{E-KAR} & \textbf{SCAN} & \textbf{Avg} \\ 
        \midrule[1pt]
        Word2Vec & 72.0 & 43.0 & 40.7 & \underline{96.6} & 47.8 & 28.2 & 16.7 & 49.3\\
        GloVe & 68.7 & 46.5 & 39.8 & 96.0 & 47.8 & 30.9 & 18.6 & 49.8\\
        FastText & 63.8 & 40.4 & 39.6 & 93.2 & 41.8 & 31.4 & 21.7 & 56.5 \\
        \midrule[1pt]
        Chen$_{BERT}$ & 68.0 & 32.8 & 34.4 & 86.6 & 38.9 & 37.9 & 14.5 & 44.7\\
        Chen$_{RoBERTa}$ & 78.2 & 46.0 & 40.0 & \textbf{96.9} & 51.6 & 46.7 & 12.1 & 53.0\\ 
        RelBERT$_{BERT}$ & 70.3 & 59.6 & 57.4 &  89.2 & 59.9 & 41.2 & 25.9 & 57.6\\
        RelBERT$_{RoBERTa}$& \underline{78.8} & \underline{66.2} & \textbf{63.0} & 92.4 & \underline{70.6} & \underline{48.8} & \underline{27.2} & \underline{63.8} \\ 
        mPLL$_{BERT}$ & 67.9 & 44.7 & 41.2 & 88.8 & 41.8 & 39.8 &  14.5 & 48.4\\
        mPLL$_{RoBERTa}$ & 78.4 &  { 58.3} & 57.4 & 93.6 &  53.4 & 44.0 & 12.1 & 56.7\\ 
        \midrule[1pt]
        Chen$^+_{RoBERTa}$ & 57.0 & 30.4 & 36.7 & 73.5 & 40.5 & 38.5 & 13.2 & 41.4 \\ 
        RelBERT$^+_{RoBERTa}$ & 53.0 & 31.2 & 35.5 & 80.5 & 44.0 & 38.8 & 18.5 & 44.6 \\ 
        \midrule[1pt]
        SememeLM$_{BERT}$ & 74.4 & 58.6 & 50.0 & 92.0 & 64.8 & 43.5 & 23.3 & 58.1 \\
        SememeLM$_{RoBERTa}$ & \textbf{82.0} & \textbf{70.8} & \underline{60.3} & 89.0 & \textbf{74.9} & \textbf{56.5} & \textbf{30.7} &  \textbf{66.3}\\ 
        \midrule[1pt]
        Human & 84.9 & 87.5 & 66.7 & 99.4 & 57.0 & 77.8 & - & -\\ 
        \bottomrule[1.5pt]
        \end{tabular}
    \label{main result}
\end{table*}

\begin{table*}[t]
\centering
\caption{We compare our method with large language models on different benchmarks. The subscript represents one-shot or zero-shot. }
        \begin{tabular}{c|cccccccc}
        \toprule[1.5pt]
        \textbf{Model}  & \textbf{BATS} & \textbf{UNIT 2} & \textbf{UNIT 4} & \textbf{Google} &\textbf{SAT} &  \textbf{E-KAR} & \textbf{SCAN} & \textbf{Avg} \\ 
        \midrule[1pt]
        InstructGPT${_0}$ & 57.7 & 47.8 & 46.9 & 78.4 & 37.3 & 32.4 &  14.5  & 45.0\\
        ChatGPT${_0}$ & 81.7 & 53.0 & 52.3 & 93.8 & 49.2 & 41.2 & 19.8 & 55.9\\
        GPT-4${_0}$ & \underline{92.4} & \underline{76.3} & \underline{71.3} & \underline{98.8} & {74.8} & {53.1} & {23.5} & \underline{70.0}\\
        InstructGPT${_1}$ & { 81.9} & 50.0 & 52.7 & 91.6 & 48.9 & 38.9 & 14.3  & 54.0\\ 
        ChatGPT${_1}$ & 81.5 & 59.2 & { 55.3} & 94.8 & 55.1 & 44.2 & 22.8 & 59.0\\ 
        GPT-4${_1}$ & \textbf{94.0} & \textbf{84.2} & \textbf{81.7} & \textbf{100.0} & \textbf{83.7} & \textbf{60.4} & \underline{26.0} & \textbf{75.7}\\ 
        \midrule[1pt]
        SememeLM$_{RoBERTa}$ & {82.0} & {70.8} & {60.3} & 89.0 & \underline{74.9} & \underline{56.5} & \textbf{30.7} & {66.3} \\  
        \bottomrule[1.5pt]
        \end{tabular}
\label{largemodels}
\end{table*}

\subsection{Experimental Setup} \label{4.1}

\subsubsection{Datasets}

\textbf{Training Data}
We employed the SemEval 2012 Task 2 dataset for our training data, which encompasses crowdsourced evaluations of 79 nuanced semantic relations categorized into 10 parent categories. The top 10 pairs, signifying those with the most elevated typicality scores, were designated as positive instances of the relation, while the lowest 10 pairs were classified as negative instances. For training purposes, 80\% of both positive and negative examples were utilized, with the remaining 20\% allocated for validation. This configuration is the same as RelBERT \cite{relbert}.\\
\textbf{Analogical Reasoning Benchmarks}
We conducted experiments on seven analogy datasets, which can be categorized into three groups. 
\begin{itemize}
    \item The first group comprises lexical semantics analogy benchmarks, including Google \cite{mikolov2013linguistic} and BATS \cite{gladkova2016analogy}. These benchmarks have been derived from human exams. 
    \item The second group consists of psychometric analogy benchmarks, which encompass E-KAR \cite{chen-etal-2022-e}, UNIT 2 \cite{BERTNLPALEXNETCV}, UNIT 4 \cite{BERTNLPALEXNETCV}, and SAT \cite{turney2003combining}. 
    For E-KAR, we used its English version, comprising a total of 1,251 analogical problems. 
    \item The third type of dataset is SCAN\cite{scan, relbert}, collected from relation mapping problem. The number of answer candidates per question is 74 on average, which makes this benchmark particularly challenging.
\end{itemize}

\begin{table*}[ht]
\centering
\caption{Ablation Study on our approach. All experimental results were completed on the basis of Roberta-large. The bolded results are the best-performing ones on this dataset.}
\begin{tabular}{@{}c|c|cccccccc@{}}
\hline
\textbf{Index} & {\textbf{Model}} & \textbf{BATS} & \textbf{UNIT 2} & \textbf{UNIT 4} & \textbf{Google} &\textbf{SAT} &  \textbf{E-KAR} & \textbf{SCAN} \\ 
\hline
1 & Ours        & \textbf{82.2} & {70.8} & \textbf{60.4} & 88.0 & \textbf{75.6} & \textbf{56.4} & {30.2} \\
2 & Ours w/o $\mathcal{L}_1$ & 78.6 & \textbf{72.6} & 53.5 & 92.6 & 53.4 & 46.6 & 24.5 \\
3 & Ours w/o $\mathcal{L}_2$ & 79.2 & 68.9 & 62.4 & 88.8 & 51.9 & 43.8 & \textbf{30.9} \\

4 & Ours w/o $\mathcal{L}_3$ & 80.2 & 46.0 & 40.0 & \textbf{96.9} & 51.6 & 46.7 & 27.2 \\
5 & Ours w/o $\mathcal{L}_1  \mathcal{L}_2$ & 76.6 & 66.6 & 59.0 & 94.2 & 70.6 & 46.8 & 25.0 \\
\hline
\end{tabular}
\label{Ablation}
\end{table*}

\subsubsection{Baselines}
We compare our approach with three distinct categories of baselines: word embedding, language models and large language models (LLMs). 

\begin{itemize}
    \item \textbf{Word Embedding} 
    In many word embedding models, including Word2Vec, GloVe and fastText, the relation between word pairs is to some extent captured by the vector difference of their embeddings. 
    \item \textbf{Language Models}
    When using language models to solve analogy questions, it is typically necessary to employ prompt engineering. For a question $(A, B)$ and each answer candidate $(C, D)$, mPPL \cite{BERTNLPALEXNETCV} construct the sentence “$A$ is to $B$ what $C$ is to $D$” and then compute the perplexity of each of these sentences, and predict the candidate that gives rise to the lower perplexity. 
    Chen\cite{chen-etal-2022-e} is to directly encode all word pairs in the question together, transforming it into a multi-class classification problem. 
    RelBert \cite{relbert} is also trained on relation similarity data, and it uses both classification and triplet loss to train the model.
    
    In Chen$^+_{RoBERTa}$ and RelBERT$^+_{RoBERTa}$, we concatenated semame information after words during training and testing. 
    \item \textbf{Large Language Models}
    OpenAI\footnote{https://openai.com/} released a commercial API to provide access to their LLMs such as InstructGPT \cite{lu2022learn}, ChatGPT and GPT-4. 
    InstructGPT is a variant of GPT-3 fine-tuned specifically for instructional content generation. 
    ChatGPT is designed to engage in natural language interactions and involving dialogues. 
    GPT-4 is a highly advanced language model designed for a wide range of natural language understanding and generation tasks. The testing method for LMs is consistent with the paper \cite{abs-2305-12660}.
\end{itemize} 

\subsubsection{Implementation Details}
Following previous works, we evaluate word analogy with accuracy. 
We use Bert-base-uncased \cite{bert} and Roberta-large \cite{roberta} as the input instance encoder. 
We set AdamW \cite{AdamW} as the optimizer and set the initial learning rate of LMs as 5e-6. 
The batch size of Bert-based-uncased and Roberta-large is set to be 16 and 32, respectively. 
Moreover, the maximum length of input sentences for LMs is 64, and sentences longer than this length will be truncated. 
The number of graphs updating steps/layers $L$ is 2. 
When using supervised contrastive loss, the temperature $\tau$ is set to be 0.5. During reasoning, we use cosine similarity to calculate the distances.

\subsection{Performance Analysis} \label{4.2}
\subsubsection{Small-sized Language Models}
We compare our model with word embedding baseline and PLMs baseline. The results are summarized in Table \ref{main result}. There are several observations drawn from different aspects: 
\begin{itemize}
 \item In the Google dataset, most word pairs are morphologically related, like "slow:slowest". However, the sememe knowledge enhancement is based on semantic relations, which may introduce noise and decrease the accuracy when solving problems. While our approach may exhibit slightly weaker performance on these particular datasets, it still remains competitive. 
 \item On psychometric analogy benchmarks, our approach narrows the gap between language models and human performance. Moreover, on lexical semantics analogy benchmarks, our approach has almost reached parity with human levels.
 \item Our approach shows potential on SCAN dataset. This might be because words from different domains are composed of similar sememes, and our model aligns with sememe representations, hence performing better across domains.
 \item In the E-KAR dataset,  query or candidates may contain three words, like "spider:web:bond", which requires the model to obtain the association between the three words. Our method encodes the relation  through sememes, which can effectively mine the relations between multiple words and improve accuracy.
 \item The result of Chen$^+_{RoBERTa}$ and RelBERT$^+_{RoBERTa}$ show that directly using semantic knowledge is not beneficial and may even reduce accuracy, which validates the necessity of our method.
\end{itemize}
Overall, our approach outperforms various word analogy approaches based on small-sized language models. This indirectly reflects our model's ability to capture the quality of relation representations.

\subsubsection{Large Language Models}
In addition to comparing with normal models, we also compared with three large language models released by OpenAI. The experimental results are shown in the table \ref{largemodels}. 
There are several observations drawn from different aspects: 
\begin{itemize} 
    \item The remarkable achievements of the LLMs were attained even without the need for fine-tuning, courtesy of their extensive training corpus and large model parameters. 
    \item It is clear that providing an example yields significantly better performance compared to not providing an example on LLMs.
    \item Our approach, after fine-tuning, surpasses InstructGPT and ChatGPT, achieving a performance level ranging from 80\% to 90\% of GPT-4. 
\end{itemize}
In general, our approach can rival typical LLMs and falls slightly short of GPT-4. To some extent, this indicates the practicality of our approach in obtain representations and addressing relation similarity problems. As GPT-4 is the most expensive endpoint at the moment, our approach is comparatively more cost-effective and scalable.

\begin{figure*}[!h]
 \centering
 \includegraphics[width=\textwidth]{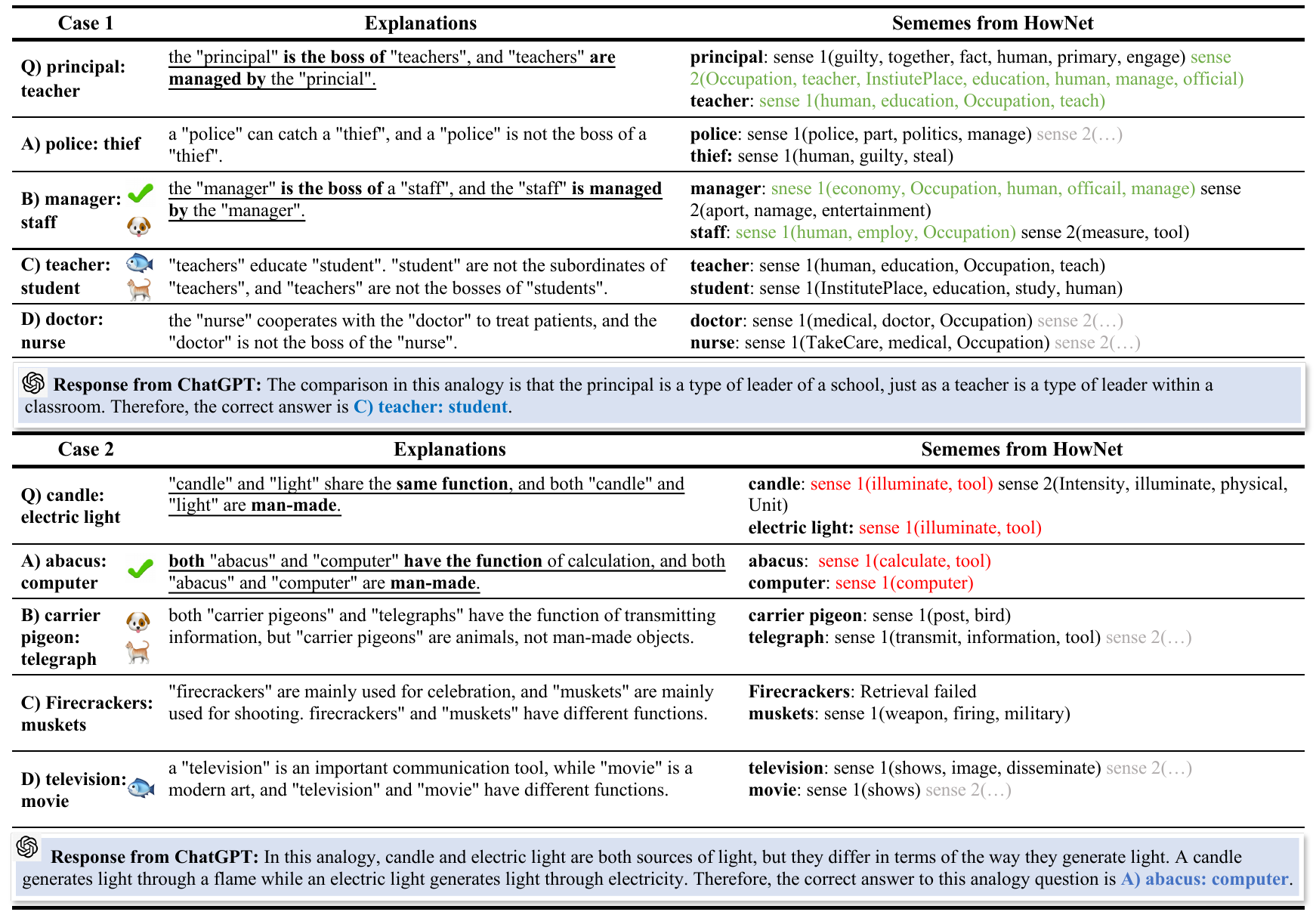}
 \caption{We analyzed the results of our approach, RelBERT, Chen$_{RoBERTa}$ and ChatGPT$_0$ on two examples in E-KAR dataset. 
 The kitten represents the choice made by Chen$_{RoBERTa}$, the fish represents the choice made by RelBERT, while the puppy illustrates the choice made by our approach.
 }
 \label{case}
\end{figure*}

\subsection{Ablation Study} \label{4.3}

To validate the effectiveness of our approach, we do ablation experiments and we performed these experiments on the Roberta-large model. The results are shown in Table \ref{Ablation}. The w/o $\mathcal{L}_1$ means removing word representation alignment. The w/o $\mathcal{L}_2$ means removing relation representation alignment. The w/o $\mathcal{L}_3$ means removing supervised contrastive loss. The w/o $\mathcal{L}_1\mathcal{L}_2$ means alignment module. 

Through ablation experiments, we found that the improvements are not synchronous across all datasets. For instance, when there is an improvement in the Google dataset, it can lead to a decrease in performance on datasets like E-KAR and SAT. 
Comparing Model 1 and Model 2, it can be found that word representation alignment is effective. Comparing Model 1 and Model 3, it can be found that relation representation alignment is effective. Our intention is to obtain relation representations from the pretrained model without disrupting the knowledge already present in the pretrained model, which is preferable. 



\subsection{Case Study} \label{4.4}
To evaluate the analogical reasoning capacity concerning long-tail relations, we extracted a some samples from the E-KAR dataset based on provided explanations. These samples required comparing some uncommon relations to solve.

Some examples are depicted in Figure \ref{case}. To comprehend how language models generate responses in the absence of linguistic knowledge, we analyzed the explanations produced by ChatGPT. Because the explanations generated by ChatGPT are relatively long, we have extracted the key portions.
In Case 1, both ChatGPT, RelBERT and PLM model opt for the most contextually relevant candidate rather than the one with the highest relation similarity. Specifically, ChatGPT focuses on a single relation and is influenced by the semantics associated with "school." In contrast, our approach performs admirably.
The sememes highlighted in green demonstrate how our approach selects the correct answer. 
Our approach, to some extent, bridges the semantic gaps between different domains, making similar relations between different domains closer in the semantic space. 

While our approach demonstrates feasibility, it does have certain limitations. It faces failure when retrieval does not yield results or when no valid information is retrieved (as seen in Case 2).
Importantly, all methods, including ours, fail to address this particular challenge, underscoring the difficulty of solving it. 

\section{Conclusion and Future Work}
In this paper, we propose a sememe knowledge enhanced method SememeLM for learning relation representation of word pairs without context. Experimental findings demonstrate that our approach outperforms some state-of-the-art methods and has shown its potential in addressing long-tail relations. 
In essence, our approach offers a means of acquiring high-quality relation representations of word pairs without context. This provides a new way for the development of related tasks such as knowledge graph completion, which could potentially inspire the constructing universal KGs by integrating language models. 


\section{Limitations}
Due to our reliance on OpenHowNet, our method is affected by database quality. It faces failure when retrieval does not yield results or when no valid information is retrieved. Our method is an idea that theoretically, it can obtain richer and more accurate relations with a well-established database.

In addition, we will conduct large-scale model comparison experiments no later than December 2024, and the experimental results may not be consistent with the constantly updated model results. We will try our best to conduct more experiments.



\bibliography{custom}

\appendix

\section{Related Work}
\subsection{Lexical Relation Representation and Analogy}
In the field of relation representation, relation induction was initially expressed through vector offsets between two entities. However, some researchers have challenged the effectiveness of this approach, leading to the exploration of alternative methods for inducing relation knowledge from pre-trained LMs \cite{bouraoui2020inducing, wang2021keml}, including KEML \cite{wang2021keml}, RI-BERT \cite{bouraoui2020inducing} and IST \cite{sun-etal-2022-minimally}. It's worth noting that while intuitively sourcing templates from a corpus is a common approach, it often introduces noise. 

Word analogy evaluations initially focused on linear word relations, often addressed effectively by neural word representation vector algorithms like Word2Vec, fastText, and GloVe. However, with the emergence of LMs, word analogy have extended to assess the performance of these models \cite{chen-etal-2022-e,relbert,DBLP:conf/emnlp/RezaeeC22,BERTNLPALEXNETCV,DBLP:journals/corr/abs-2305-12660}. Such evaluations typically employ prompts, either manually crafted or automatically generated, to encode queries and candidates. 

\subsection{Knowledge Enhanced Language Model}
Currently, most methods for acquiring long-tail relations are primarily focused on relation extraction \cite{DBLP:journals/tkde/CaoKGZWC23, DBLP:journals/tkde/LiangLLZSG23, DBLP:conf/emnlp/WanLMCKL22}. 
In addition to relation extraction, there are also numerous methods for enhancing pre-trained models with knowledge graphs.
KnowBERT \cite{DBLP:conf/emnlp/PetersNLSJSS19} achieves this by incorporating knowledge bases into BERT using Knowledge Attention and Recontextualization, with knowledge derived from synset and lemma relations in WordNet. 
K-adapter \cite{DBLP:conf/acl/WangTDWHJCJZ21} adds an adapter to infuse factual knowledge into LMs without modifying the original parameters. 
Furthermore, LET \cite{DBLP:conf/aaai/LyuCZ021} introduces HowNet as an external knowledge base for addressing Chinese short text matching tasks, emphasizing semantic levels. 

\subsection{Inducing Knowledge from LMs}
Recently, exploratory tasks have been employed to gain a deeper insight into the underlying nature of the representations acquired by neural language models, though most studies have predominantly centered on factual and linguistic facets \cite{bouraoui2020inducing,relbert,MultilingualKnowledgeTriples2022emnlp}. In a more closely related context to our research, RI-BERT \cite{bouraoui2020inducing} automatically discern the most fitting trigger sentences for each relation and concentrate on relation classification. KEML \cite{wang2021keml} develops a classifier to predict the lexical relation between two entities based on relation labels. Similarly, RelBERT \cite{relbert} encodes word pairs via a prompt and refines the LM during fine-tuning, thereby rendering relationally analogous word pairs with akin vectors. During model training, they also fine-tune the LM. 

\end{document}